\documentclass[preprint,12pt]{elsarticle}

\usepackage{amssymb}
\usepackage{multirow}
\usepackage{graphicx}
\usepackage{array, booktabs, longtable}	
\usepackage{lineno,hyperref}
\usepackage{multirow}

\journal{VSI: AI in Breast Cancer Care}

\begin{document}

\begin{frontmatter}



\title{Multi-class Semantic Segmentation of Skin Lesions via Fully Convolutional Networks}

\author[label1]{Manu Goyal}
\author[label2]{Moi Hoon Yap}
\author[label3]{Saeed Hassanpour}
\address[label1]{Department of Biomedical Data Science, Dartmouth College, Hanover, NH, USA.}
\address[label2]{Visual Computing Lab, Manchester Metropolitan University, Manchester, UK.}
\address[label3]{Departments of Biomedical Data Science, Computer Science, and Epidemiology, Dartmouth College, Hanover, NH, USA.}

\address{}

\begin{abstract}
Melanoma is clinically difficult to distinguish from common benign skin lesions, particularly melanocytic naevus and seborrhoeic keratosis. The dermoscopic appearance of these lesions has huge intra-class variations and high inter-class visual similarities. Most current research is focusing on single-class segmentation irrespective of classes of skin lesions. In this work, we evaluate the performance of deep learning on multi-class segmentation of ISIC-2017 challenge dataset, which consists of 2,750 dermoscopic images. We propose an end-to-end solution using fully convolutional networks (FCNs) for multi-class semantic segmentation to automatically segment the melanoma, seborrhoeic keratosis and naevus. To improve the performance of FCNs, transfer learning and a hybrid loss function are used. We evaluate the performance of the deep learning segmentation methods for multi-class segmentation and lesion diagnosis (with post-processing method) on the testing set of the ISIC-2017 challenge dataset.  The results showed that the two-tier level transfer learning FCN-8s achieved the overall best result with \textit{Dice} score of 78.5\% in a naevus category, 65.3\% in melanoma, and 55.7\% in seborrhoeic keratosis in multi-class segmentation and \textit{Accuracy} of 84.62\% for recognition of melanoma in lesion diagnosis.

\end{abstract}




\end{frontmatter}

\section{Introduction}

Skin cancers are more common than all other cancers \cite{pathan2018techniques}. Malignant skin lesions are classified as melanocytic, i.e. melanoma, and non-melanocytic. The most common non-melanocytic cancers are keratinocytic: basal cell carcinoma and squamous cell carcinoma. Melanoma is less common but is more likely to prove fatal than keratinocytic skin cancers due to aggressive invasion and metastasis \cite{Seer2017}\cite{dvovrankova2017intercellular}. Hence, early detection is important to save lives. According to the prediction of the Melanoma Foundation \cite{melanomafoundation2017}, the estimated diagnosed cases of melanoma in the United States in 2018 is 178,560 with 91,270 cases will be invasive. 

Melanocytic naevi and seborrhoeic keratosis are very common benign skin lesions that may be clinically difficult to differentiate from skin cancer. Both melanoma and melanocytic naevi are melanocytic lesion as uncontrolled growth of melanocytes (pigmented cells) results in melanoma whereas non-cancerous growth in moles results in benign melanocytic naevus. Seborrheic keratosis is a type of non-melanocytic skin lesion. But, it is very hard to distinguish the SK lesions from melanocytic lesions (moles and melanoma) even with the help of dermoscopy as these skin lesions share similar features such as irregular shapes and multiple colors.

With the rapid growth of deep learning approaches, many researchers \cite{yuan2017automatic}, \cite{yu2017automated}, \cite{bi2017dermoscopic}, \cite{8936444} have proposed Deep Convolutional Neural Networks for skin lesion segmentation (single-class). We have found no previous research on multi-class semantic segmentation for different types of skin lesions. 

Our contributions are three fold. Firstly, we propose multi-class semantic lesions segmentation for melanoma, seborrhoeic keratosis and naevus. To overcome data deficiency, a two-tier transfer learning is used in skin lesions segmentation to train the fully convolutional networks (FCNs). Secondly, we design a hybrid loss function to handle class imbalance in the multi-class segmentation. Thirdly, we assess the performance of state-of-the-art deep learning algorithms using our proposed multi-class segmentation and a post-processing method to determine lesion diagnosis  on ISIC-2017 Challenge dataset. Our proposed method can be generalised into other multi-class segmentation tasks in medical imaging.

\section{Methodology}
This section discusses the publicly available ISIC-2017 skin lesion dataset and its ground truth labeling, the two-tier transfer learning approach, and the hybrid loss function.

\subsection{Datasets and Ground Truth}
We used the publicly available ISIC-2017 \textit{Skin Lesion Analysis Towards Melanoma Detection Challenge} dataset \cite{codella2017skin} to train the fully convolutional deep learning models. RGB colorspace is used to represent all the images in this dataset. It includes 3 skin lesion types using dermoscopy images: naevi, melanomas and seborrhoeic keratosis. The segmentation task on these dermoscopy images is very challenging due to high inter-class similarity between the 3 types of skin lesions. This dataset is imbalanced as there are only a total of 521 melanoma and 386 seborrheic keratosis compared to 1843 melanocytic naevi dermoscopic images. There are a total of 2750 dermoscopy images in the ISIC-2017 challenge dataset, as summarised in Table \ref{tab:tradFse}. 


\begin{table}[]
	\centering
	\small\addtolength{\tabcolsep}{2pt}
	\renewcommand{\arraystretch}{2}\vspace{0.25cm}
	\caption{Distribution of images for multi-class segmentation task. }
	\label{tab:tradFse}
	\scalebox{0.7}{
		\begin{tabular}{ccccc}
			\hline
			& Naevi & Melanoma & Seborrheic Keratosis & Total \\ \hline\hline
			Training set   & 1372        & 521      & 387                   & 2000  \\
			Validation set & 92          & 34       & 23                   & 150   \\
			Testing set       & 393         & 117      & 90                    & 600   \\
			Total           & 1843        & 521      & 386                   & 2750 \\ \hline
	\end{tabular}}
\end{table}

\begin{figure*}[ht]
	\centering
	\begin{tabular}{ccc}
		\includegraphics[width=2.6cm,height=2.1cm]{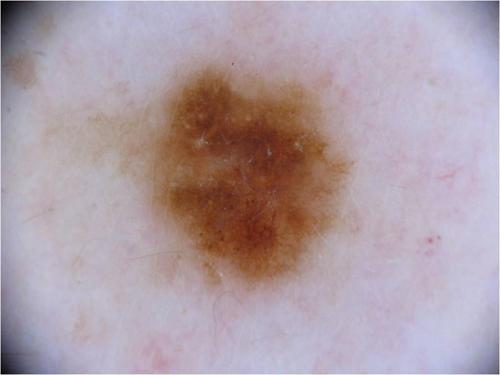} &
		\includegraphics[width=2.6cm,height=2.1cm]{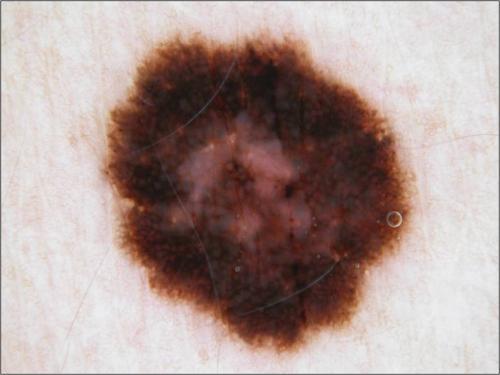} &		
		\includegraphics[width=2.6cm,height=2.1cm]{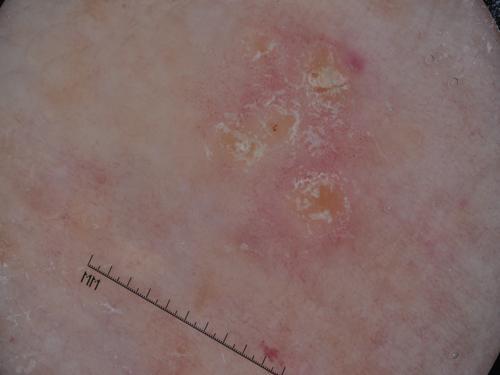} \\
		\includegraphics[width=2.6cm,height=2.1cm]{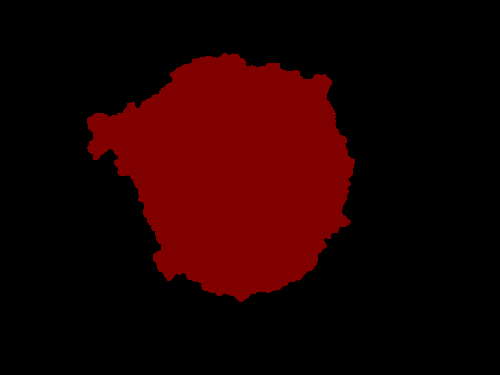} &
		\includegraphics[width=2.6cm,height=2.1cm]{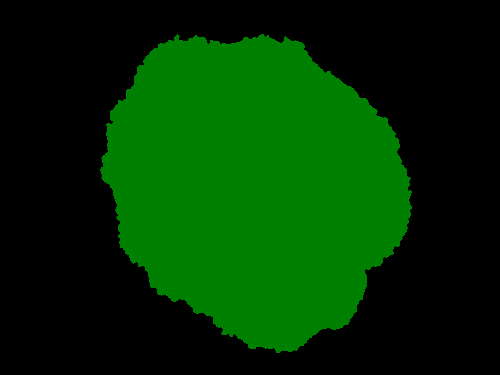} &
		\includegraphics[width=2.6cm,height=2.1cm]{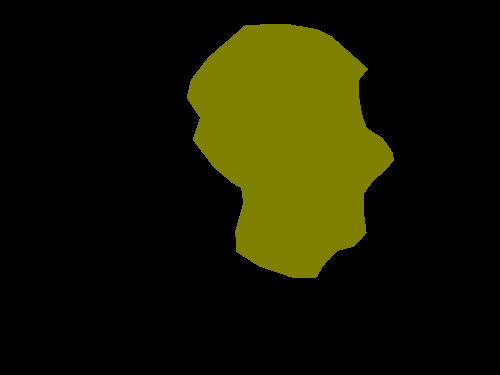} \\
		(a) & (b)&(c) \\
	\end{tabular}
	\caption[The sample 8-bit paletted label images.]{Original images (first row) and PASCAL-VOC format (second row). The skin lesion diagnosis from left to right: (a) naevus, (b) melanoma and (c) seborrhoeic keratosis.}
	\label{fig:datasetVOC}
\end{figure*}

In this dataset, the size of images varies between 540 $\times$ 722 and 4499 $\times$ 6748. To improve the performance and reduce the computational cost, all the images are resized to 500 $\times$ 375. In ISIC-2017 segmentation challenge, the task is to segment the lesion boundaries, which was a one-class segmentation task. Here we are targeting on automatic multi-class segmentation. The ground truths are all defined in RGB colorspace and 8-bit paletted images. Figure \ref{fig:datasetVOC} illustrates the dermoscopic images with the corresponding ground truth labeling in PASCAL-VOC format  \cite{garcia2017review}\cite{Everingham15}. Index 1 indicates naevus, index 2 indicates melanoma and index 3 represents seborrhoeic keratosis. 

\subsection{Fully Convolutional Networks for Multi-class Semantic Segmentation}
FCNs and encoder-decoder CNNs can detect the multiple objects as well as localize the objects by using pixel-wise prediction. This enables to learn which pixel of an image belongs to which class of object. Recently, FCNs have become the state-of-the-art methods for segmentation tasks on both non-medical and medical imaging, which are superior to conventional machine learning and other deep learning methods. We used the four different variants of FCNs (FCN-AlexNet, FCN-32s, FCN-16s, and FCN-8s) and assessed their performance on multi-class skin lesions segmentation. 

The first variant FCN-AlexNet is a modified version of original state-of-the-art classification model called AlexNet, which won ImageNet ILSVRC-2012 competition in the classification category \cite{long2015fully}\cite{krizhevsky2012imagenet}. The FCN-AlexNet enables the pixel-wise prediction by using the deconvolutional layers which up-sample the features learned by the earlier convolutional layers. We have trained the FCN-AlexNet on the Caffe deep learning framework \cite{jia2014caffe}. The input and ground truth images are both 500$\times$375. We have fine-tuned the network parameters to allow the method more time to learn the features from dermoscopy images by using 100 epochs, stochastic gradient descent with a learning rate of 0.0001.    

The other FCNs variants, FCN-32s, FCN-16s and FCN-8s, are based on another state-of-the-art classification network called VGG-16, which won the localization challenge and was in second position for the classification challenge in the ImageNet ILSVRC-2014 competition \cite{simonyan2014very}\cite{long2015fully}. The differences between these models are the up-sampling layers with different pixel stride. As the name suggested by these FCNs variants, in FCN-32s, up-sampling is performed with the help of 32-pixel stride whereas 16-pixel stride is used for FCN-16s and 8-pixel stride for FCN-8s. With the small pixel stride, the models were able to predict finer-grained analysis of the objects. The same network parameters as FCN-AlexNet were used to train these models.

\begin{figure*}
	\centering
	\begin{tabular}{ccc}
		\includegraphics[width=3.5cm,height=3cm]{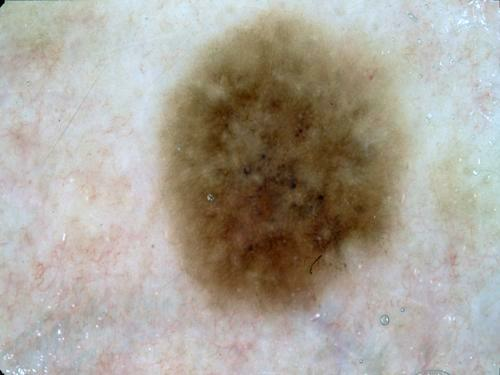} &
		\includegraphics[width=3.5cm,height=3cm]{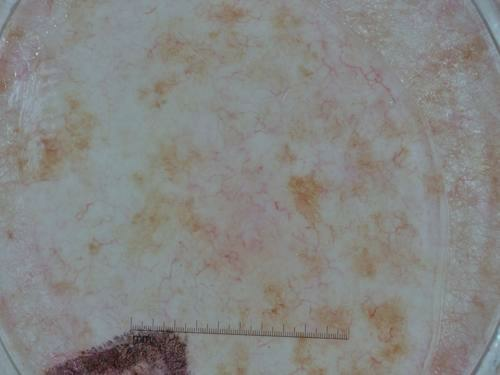} &
		\includegraphics[width=3.5cm,height=3cm]{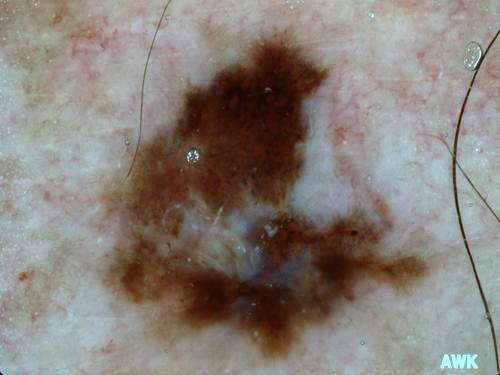} \\
		\includegraphics[width=3.5cm,height=3cm]{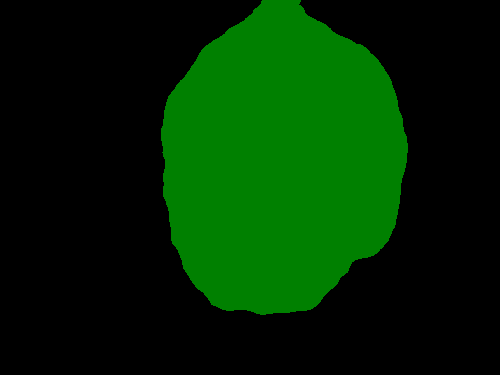} &   
		\includegraphics[width=3.5cm,height=3cm]{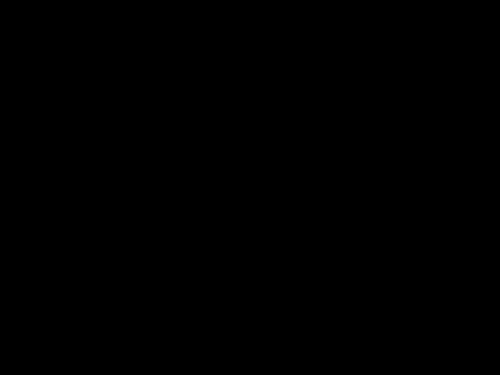}&
		\includegraphics[width=3.5cm,height=3cm]{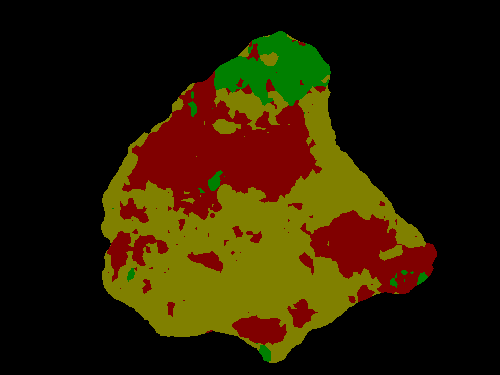}\\
		
		(a) Single-detection & (b) No-detection  & (c) Multi-detection  \\ 
	\end{tabular}     
	
	\caption[]{Examples of different types of semantic segmentation in ISIC-2017 testing set: (a) result with one class lesion type; (b) result with no lesion detected; and (c) result with multiple lesion types. Where green color represents melanocytic naevus, red color represents melanoma, and yellow color represents seborrhoeic keratosis}
	\label{fig:det1}
\end{figure*}

\begin{table}[]
	\centering
	\addtolength{\tabcolsep}{2pt}\vspace{0.25cm}
	\renewcommand{\arraystretch}{2.5}
	\caption{Number of cases for each type of inference in ISIC-2017 Testing Set}
	\label{prediction}
	\scalebox{0.70}{
		\begin{tabular}{cccc}
			\hline
			Inference      & Single-detection & Multi-detection & No-detection \\ \hline \hline
			Testing Set  &395 &192    & 13 \\ \hline        
	\end{tabular}}
\end{table}  

\begin{table*}[]
	\centering
	\normalfont\addtolength{\tabcolsep}{1pt}
	\renewcommand{\arraystretch}{2}\vspace{0.25cm}
	\caption{Comparison of different FCN architectures using the ISIC-2017 Challenge Dataset (Mel denotes Melanoma and SK denotes Seborrheic Keratosis)}
	\label{tab:tradFeats}
	\scalebox{0.6}{
		\begin{tabular}{|c|c|c|c|c|c|c|c|c|c|c|c|c|c}
			\cline{1-13}
			\multirow{2}{*}{Method} & \multicolumn{3}{c|}{\textit{Dice}}                              & \multicolumn{3}{c|}{\textit{Specificity}}                       & \multicolumn{3}{c|}{\textit{Sensitivity}}                       & \multicolumn{3}{c|}{\textit{MCC}}                               &  \\ \cline{2-13}
			& Naevi & Mel & SK    & Naevi    & Mel   & SK      & Naevi    & Mel   & SK      & Naevi & Mel & SK     &  \\ \cline{1-13}
			FCN-AlexNet             & \textbf{0.819}  & 0.609    & 0.488 & 0.989     & 0.982      & 0.987   & \textbf{0.798}     & 0.4864     & 0.456   & \textbf{0.814}  & 0.541    & 0.484  &  \\ \cline{1-13}
			FCN-32s                 & 0.779  & 0.549    & 0.484 & \textbf{0.991}     & 0.977      & 0.968   & 0.751     & 0.430      & 0.478   & 0.775  & 0.484    & 0.463  &  \\ \cline{1-13}
			FCN-16s                 & 0.761  & 0.590    & 0.506 & 0.988     & 0.979      & 0.978   & 0.706     & 0.471      & 0.466   & 0.764  & 0.528    & 0.501  &  \\ \cline{1-13}
			FCN-8s                  & 0.785  & \textbf{0.653}    & \textbf{0.557} & 0.990     & \textbf{0.984}      & \textbf{0.988}   & 0.747     & \textbf{0.527}      & \textbf{0.509}   & 0.779  & \textbf{0.582}    & \textbf{0.5683} &  \\ \cline{1-13}
	\end{tabular}}
\end{table*}

\begin{figure*}[!t]
	\centering
	\begin{tabular}{ccccccc}
		Ground truth &FCN-AlexNet&FCN-32s &FCN-16s &FCN-8s\\
		
		\includegraphics[width=2.25cm,height=2.25cm]{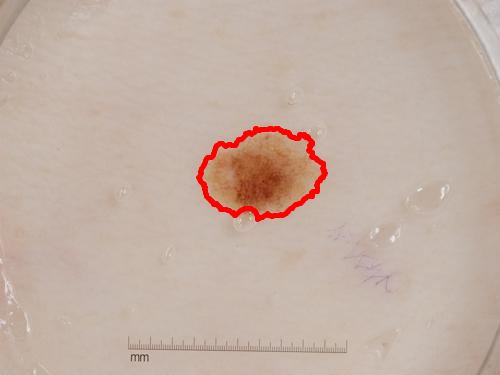} &
		\includegraphics[width=2.25cm,height=2.25cm]{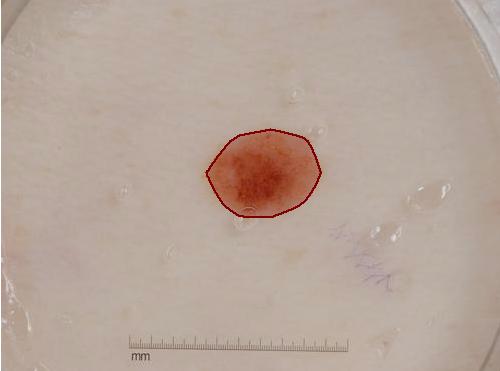}&
		\includegraphics[width=2.25cm,height=2.25cm]{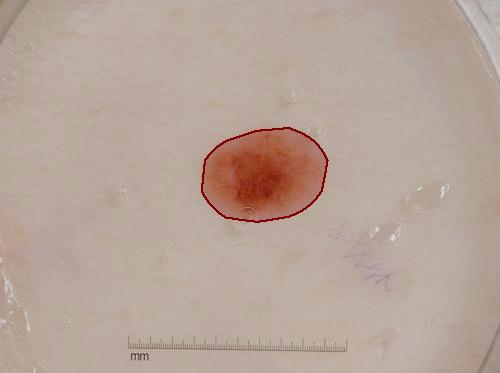}&
		\includegraphics[width=2.25cm,height=2.25cm]{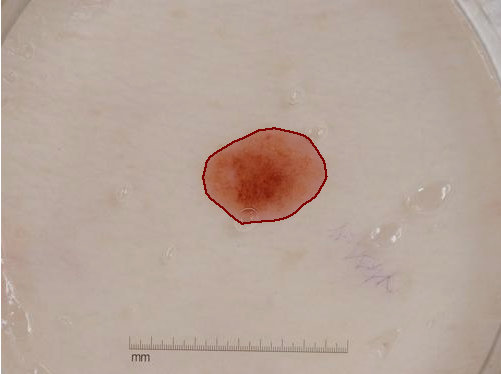}&
		\includegraphics[width=2.25cm,height=2.25cm]{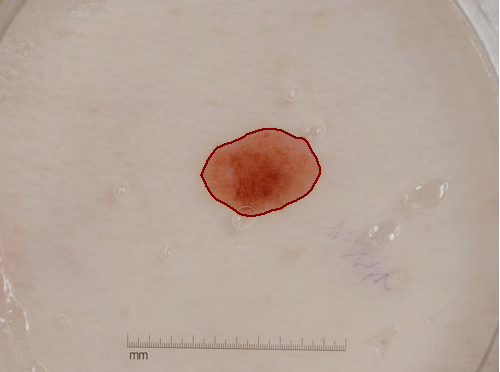}&		
		\\ 
		
		\includegraphics[width=2.25cm,height=2.25cm]{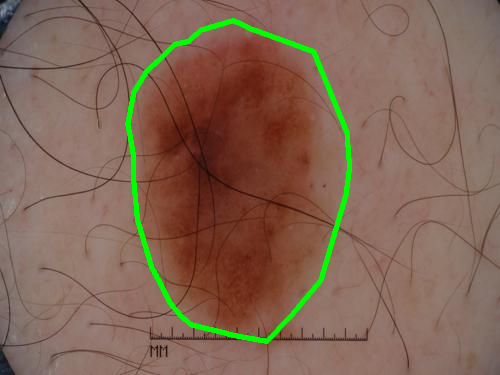} &
		\includegraphics[width=2.25cm,height=2.25cm]{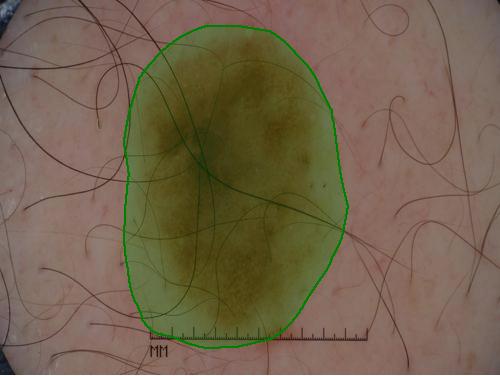}&
		\includegraphics[width=2.25cm,height=2.25cm]{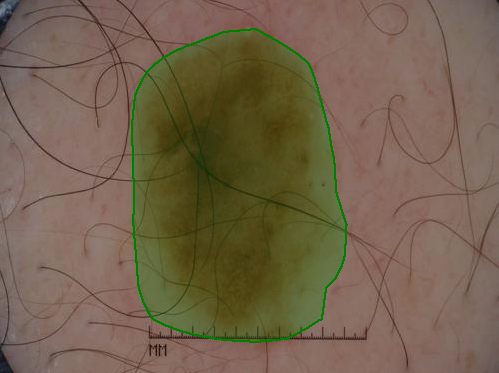}&
		\includegraphics[width=2.25cm,height=2.25cm]{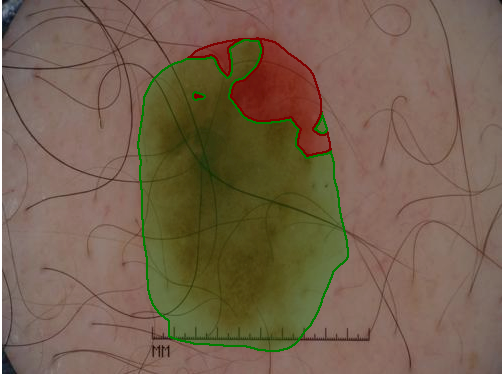}&
		\includegraphics[width=2.25cm,height=2.25cm]{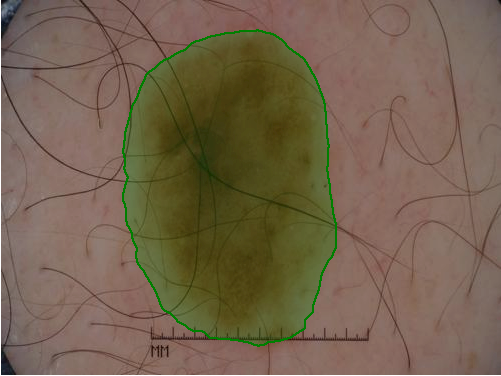}&
		\\ 
		
		\includegraphics[width=2.25cm,height=2.25cm]{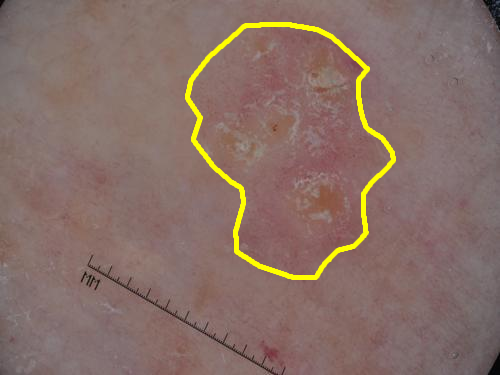} &
		\includegraphics[width=2.25cm,height=2.25cm]{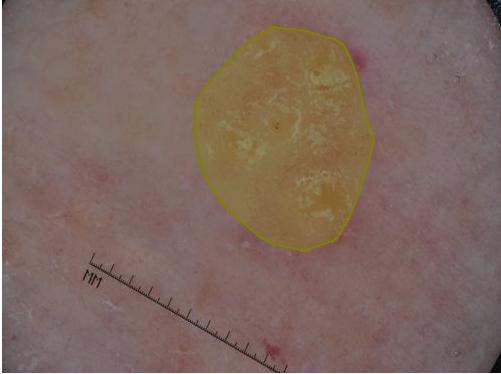}&
		\includegraphics[width=2.25cm,height=2.25cm]{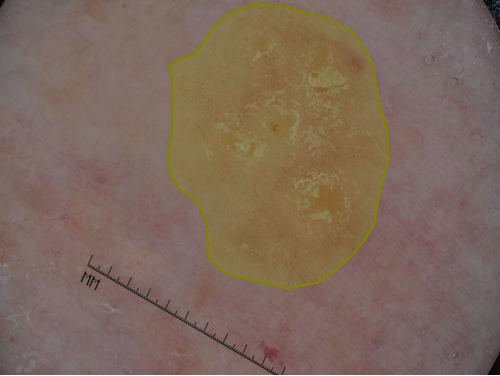}&
		\includegraphics[width=2.25cm,height=2.25cm]{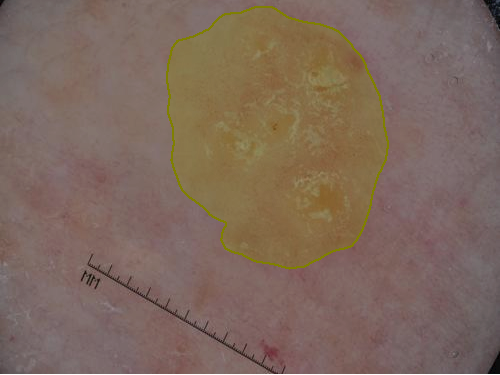}&
		\includegraphics[width=2.25cm,height=2.25cm]{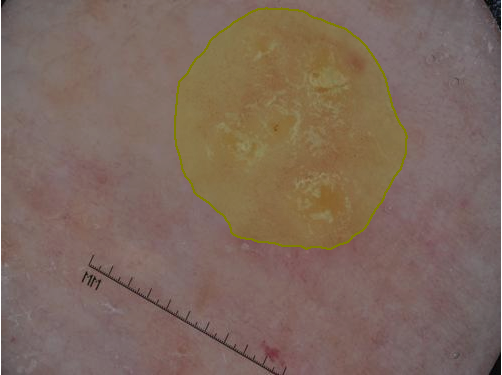}&
		
		\\ 
	\end{tabular}     
	
	\caption[Illustration of segmentation results to visually compare the performance of FCNs on multi-class segmentation.]{Illustration of segmentation results to visually compare the performance of ground truth delineation and four FCNs on multi-class segmentation for a naevus (top row), a melanoma (middle row), and a seborrhoeic keratosis (bottom row). }
	\label{fig:resultsVisual}
\end{figure*}

\subsection{The Two-tier Transfer Learning Approach}
Convolutional neural networks generally require a huge dataset to learn the features and detect objects in images \cite{lecun2015deep}. Since, we have RGB images in dermoscopic images, it is good to use two-tier transfer learning from huge datasets in non-medical backgrounds such as ImageNet and Pascal-VOC dataset to converge the weights associated with each convolutional layer of networks  \cite{russakovsky2014imagenet}\cite{Everingham15}\cite{goyal2017fully}.   
The transfer learning transfers the feature learned by previous models on huge non-medical datasets to medical image datasets. There are two types of transfer learning, i.e. partial transfer learning in which only the features from few convolutional layers are transferred, and full transfer learning in which features are transferred from all the layers of previous pre- trained models. For the first tier of two-tier transfer learning, we used partial transfer learning by transferring the features from the convolutional layers trained on ImageNet. For the second tier, we used full transfer learning from a model trained on Pascal-VOC.

\subsection{Custom Hybrid Loss function}
For imbalanced dataset as summarized in Table \ref{tab:tradFse}, we used a hybrid loss function, which is a combination of softmax cross-entropy loss and \textit{Dice} score loss function, to optimize the objective function. \textit{Dice} Score is a commonly used performance metric in medical imaging segmentation. Softmax cross-entropy loss function is a sum of per-pixel softmax cross-entropy loss whereas \textit{Dice} score loss function take care of overall segmentation score on whole image.
\begin{equation} 
L_s  = Softmax(cross-entropy)
\end{equation}	
where L\textsubscript{s} is overall softmax cross entropy loss function and cross-entropy is per-pixel cross-entropy loss.

\begin{equation} 
L_d= \frac{2|S \cap G|}{|S|+|G|}
\end{equation}
where L\textsubscript{d} is \textit{Dice} score loss function, S is segmented image and G is ground truth.

\begin{equation} 
L_h= L_s + L_d
\end{equation}
where L\textsubscript{h} is a hybrid loss function which is combination of both softmax cross entropy loss function and dice loss function.

\section{Result and Discussion}
We experimented with four state-of-the-art fully convolutional networks for our proposed segmentation task as described above. We trained the models on the ISIC-2017 training set of 2000 dermoscopic images with an input-size of 500$\times$375 using stochastic gradient descent with a learning rate of 0.0001, 60 epochs with a dropout rate of 33\%. In Table \ref{tab:tradFeats}, we report \textit{Dice Similarity Coefficient (Dice)}, \textit{Sensitivity}, \textit{Specificity}, \textit{Matthews Correlation Coefficient (MCC)} as metrics for performance evaluation of multi-class segmentation of skin lesions. We used the trained model based on the best \textit{Dice} score on the ISIC-2017 validation set to perform inference on the ISIC-2017 test set.   

\textbf{Configuration of GPU Machine for Experiments}
(1) Hardware: CPU - Intel i7-6700 @ 4.00Ghz, GPU - NVIDIA TITAN X 12Gb, RAM - 32GB DDR5 (2) Software: Caffe.

In performance measure for multi-class segmentation, we received three types of results from the inference as shown in the Fig. \ref{fig:det1} and number of cases for each type of detection is shown in Table \ref{prediction}. In Table \ref{tab:tradFeats}, we report the performance evaluation of fully convolutional networks for multi-class segmentation on ISIC-2017 test set. In the naevi category, all FCNs achieved good segmentation results, but FCN-AlexNet achieved the best results with \textit{Dice} score of 0.819, \textit{MCC} score of 0.814, and \textit{Sensitivity} is 0.798. In this category, FCN-8s performed 2nd best with \textit{Dice} score of 0.779 and \textit{MCC} score of 0.779. In the melanoma and seborrhoeic keratosis catergories, FCN-8s has achieved \textit{Dice} score of 0.653 and 0.557 respectively, which was also the best performer for all the metrics. Fig. \ref{fig:resultsVisual} visually compares the segmentation results on different lesion types. FCNs performed best in the class of naevi because we have more images of naevi than melanoma and seborrhoeic keratosis. Due to high intra-class and inter-class visual similarities, performance for both melanoma and seborrhoeic keratosis suffer due to fewer images in the dataset. Melanoma images are approx. 37\% and keratosis images are approx. 22\% of the total of images of naevi in the dataset.

The results demonstrated that deep learning techniques are reliant on the size of dataset. The segmentation results for melanoma and seborrhoeic keratosis were notably poorer than for naevi as a consequence of data deficiency. Despite the limitation on dataset, we have provided a fully automated end-to-end solution for multi-class segmentation.

\subsection{Post-processing Method to Determine Lesion Diagnosis}
We used a post-processing method to determine a single label for lesion diagnosis especially for multi-detection. We only used FCN-8s for this stage as it provided best scores for the segmentation of melanoma and seborrhoeic keratosis. For single-detection, we directly assumed the detected lesion class as same. There were very few cases of no detection (13 cases out of 600) in testing set, we assumed these cases as naevi for performance evaluation. For multi-class detection, we adopted an priority based strategy for class prediction with preference of the malignant lesions over the benign and number of images in the training set according to the Table \ref{tab:tradFseddd}. For example, the (c) multi-detection case in Fig. \ref{fig:det1} is classified as melanoma according to priority based strategy.

\begin{table}[]
	\centering
	\small\addtolength{\tabcolsep}{2pt}\vspace{0.25cm}
	\renewcommand{\arraystretch}{2}
	\caption{Priority strategy based on benign/malignant and number of images in ISIC-2017 training set. Where SK is seborrhoeic keratosis}
	\label{tab:tradFseddd}
	\scalebox{0.78}{
		\begin{tabular}{lllc}
			\hline
			Priority & Class & Benign/Malignant & No. of Images   \\ \hline\hline
			1 & Melanoma       & Malignant & 541\\ 
			2 & SK  & Benign & 387                  \\
			3 & Naevi         & Benign & 1372 \\ \hline
	\end{tabular}}
\end{table}

In Table \ref{prediction1}, we report the performance of selected FCN-8s with post-processing method to determine lesion diagnosis. We achieved an \textit{Accuracy} of 84.62\% for recognition of melanoma and 74.44\% for seborrhoeic keratosis with our proposed post-processing method despite the poor performance of FCNs for segmentation of melanoma and seborrhoeic keratosis.

\begin{table}[!t]
	\centering
	\addtolength{\tabcolsep}{2pt}\vspace{0.2cm}
	\renewcommand{\arraystretch}{1.5}
	\caption{The performance of FCN-8s with post-processing method for lesion diagnosis on ISIC-2017 testing set. Where SK is seborrhoeic keratosis}
	\label{prediction1}
	\scalebox{0.7}{
		\begin{tabular}{c c c c c} 
			\hline 
			Class   & No. of Cases & Correct & Incorrect & Accuracy\\
			\hline\hline
			
			Naevi & 393&319&74&81.17\\
			Melanoma    & 117 & 99&18&84.62\\
			SK  & 90 & 67&23&74.44\\ 
			Overall  & 600 & 485&115&80.83\\ \hline

	\end{tabular}}
\end{table}

\section{Conclusion}
We propose a fully automated multi-class semantic segmentation for melanomas, naevi and seborrhoeic keratosis in the ISIC 2017 Challenge dataset. Segmentation of skin lesions is very challenging as there are high intra-class variations and inter-class similarities in terms of visual appearance, size and colour. The literature on skin lesion segmentation only describes one-class solutions. Computer vision algorithms can easily segment one class of skin lesion from the surrounding healthy skin. But it remains a major challenge to achieve good multi-class segmentation results for multiple categories. We designed a hybrid loss function and implemented two-tier transfer learning and successfully established a new baseline for multi-class segmentation for skin lesions. We further investigated the post-processing method to improve the lesion diagnosis of FCNs. With balanced skin lesion dataset and expert annotation, the method has potential to further improve the lesion diagnosis with multi-class segmentation.



\section*{Reference}
 \bibliographystyle{elsarticle-num} 
 \bibliography{skin.bib}

\begin{thebibliography}{10}
\expandafter\ifx\csname url\endcsname\relax
  \def\url#1{\texttt{#1}}\fi
\expandafter\ifx\csname urlprefix\endcsname\relax\def\urlprefix{URL }\fi
\expandafter\ifx\csname href\endcsname\relax
  \def\href#1#2{#2} \def\path#1{#1}\fi

\bibitem{pathan2018techniques}
S.~Pathan, K.~G. Prabhu, P.~Siddalingaswamy, Techniques and algorithms for
  computer aided diagnosis of pigmented skin lesions—a review, Biomedical
  Signal Processing and Control 39 (2018) 237--262.

\bibitem{Seer2017}
{National Cancer Institute},
  \href{https://seer.cancer.gov/statfacts/html/melan.html}{Cancer stat facts:
  Melanoma of the skin}, last access: 26/10/17 (2017).
\newline\urlprefix\url{https://seer.cancer.gov/statfacts/html/melan.html}

\bibitem{dvovrankova2017intercellular}
B.~Dvo{\v{r}}{\'a}nkov{\'a}, P.~Szabo, O.~Kodet, H.~Strnad, M.~Kol{\'a}{\v{r}},
  L.~Lacina, E.~Krej{\v{c}}{\'\i}, O.~Na{\v{n}}ka, A.~{\v{S}}edo, K.~Smetana,
  Intercellular crosstalk in human malignant melanoma, Protoplasma (2017) 1--8.

\bibitem{melanomafoundation2017}
{Melanoma Foundation (AIM)},
  \href{https://www.aimatmelanoma.org/about-melanoma/melanoma-stats-facts-and-figures/}{Melanoma
  stats, facts and figures}, last access: 27/10/2017 (2017).
\newline\urlprefix\url{https://www.aimatmelanoma.org/about-melanoma/melanoma-stats-facts-and-figures/}

\bibitem{yuan2017automatic}
Y.~Yuan, M.~Chao, Y.-C. Lo, Automatic skin lesion segmentation using deep fully
  convolutional networks with jaccard distance, IEEE Transactions on Medical
  Imaging.

\bibitem{yu2017automated}
L.~Yu, H.~Chen, Q.~Dou, J.~Qin, P.-A. Heng, Automated melanoma recognition in
  dermoscopy images via very deep residual networks, IEEE transactions on
  medical imaging 36~(4) (2017) 994--1004.

\bibitem{bi2017dermoscopic}
L.~Bi, J.~Kim, E.~Ahn, A.~Kumar, M.~Fulham, D.~Feng, Dermoscopic image
  segmentation via multi-stage fully convolutional networks, IEEE Transactions
  on Biomedical Engineering.

\bibitem{8936444}
M.~{Goyal}, A.~{Oakley}, P.~{Bansal}, D.~{Dancey}, M.~H. {Yap}, Skin lesion
  segmentation in dermoscopic images with ensemble deep learning methods, IEEE
  Access (2019) 1--1\href {http://dx.doi.org/10.1109/ACCESS.2019.2960504}
  {\path{doi:10.1109/ACCESS.2019.2960504}}.

\bibitem{codella2017skin}
N.~C. Codella, D.~Gutman, M.~E. Celebi, B.~Helba, M.~A. Marchetti, S.~W. Dusza,
  A.~Kalloo, K.~Liopyris, N.~Mishra, H.~Kittler, et~al., Skin lesion analysis
  toward melanoma detection: A challenge at the 2017 international symposium on
  biomedical imaging (isbi), hosted by the international skin imaging
  collaboration (isic), arXiv preprint arXiv:1710.05006.

\bibitem{garcia2017review}
A.~Garcia-Garcia, S.~Orts-Escolano, S.~Oprea, V.~Villena-Martinez,
  J.~Garcia-Rodriguez, A review on deep learning techniques applied to semantic
  segmentation, arXiv preprint arXiv:1704.06857.

\bibitem{Everingham15}
M.~Everingham, S.~M.~A. Eslami, L.~Van~Gool, C.~K.~I. Williams, J.~Winn,
  A.~Zisserman, The pascal visual object classes challenge: A retrospective,
  International Journal of Computer Vision 111~(1) (2015) 98--136.

\bibitem{long2015fully}
J.~Long, E.~Shelhamer, T.~Darrell, Fully convolutional networks for semantic
  segmentation, in: Proceedings of the IEEE Conference on Computer Vision and
  Pattern Recognition, 2015, pp. 3431--3440.

\bibitem{krizhevsky2012imagenet}
A.~Krizhevsky, I.~Sutskever, G.~E. Hinton, Imagenet classification with deep
  convolutional neural networks, in: Advances in neural information processing
  systems, 2012, pp. 1097--1105.

\bibitem{jia2014caffe}
Y.~Jia, E.~Shelhamer, J.~Donahue, S.~Karayev, J.~Long, R.~Girshick,
  S.~Guadarrama, T.~Darrell, Caffe: Convolutional architecture for fast feature
  embedding, in: Proceedings of the 22nd ACM international conference on
  Multimedia, ACM, 2014, pp. 675--678.

\bibitem{simonyan2014very}
K.~Simonyan, A.~Zisserman, Very deep convolutional networks for large-scale
  image recognition, arXiv preprint arXiv:1409.1556.

\bibitem{lecun2015deep}
Y.~LeCun, Y.~Bengio, G.~Hinton, Deep learning, Nature 521~(7553) (2015)
  436--444.

\bibitem{russakovsky2014imagenet}
O.~Russakovsky, J.~Deng, H.~Su, J.~Krause, S.~Satheesh, S.~Ma, Z.~Huang,
  A.~Karpathy, A.~Khosla, M.~Bernstein, et~al., Imagenet large scale visual
  recognition challenge, arXiv preprint arXiv:1409.0575.

\bibitem{goyal2017fully}
M.~Goyal, N.~D. Reeves, S.~Rajbhandari, J.~Spragg, M.~H. Yap, Fully
  convolutional networks for diabetic foot ulcer segmentation, arXiv preprint
  arXiv:1708.01928.

\end{thebibliography}





\end{document}